\documentclass[conference]{IEEEtran}
\IEEEoverridecommandlockouts
\def\IEEEtitletopspace{18pt}
\usepackage{amsmath,amssymb,amsfonts}
\usepackage{algorithmic}
\usepackage{graphicx}
\usepackage{textcomp}
\usepackage{xcolor}
\usepackage{subcaption}
\usepackage{caption}

\usepackage{hyperref}
\usepackage[capitalise]{cleveref}

\usepackage[sorting=none]{biblatex}
\usepackage{enumitem}
\setitemize{noitemsep,topsep=0pt,parsep=0pt,partopsep=0pt}
\usepackage{multicol}
\usepackage{booktabs}

\begin{document}

\title{Gotta catch 'em all, safely! Aerial-deployed soft underwater gripper\\

\thanks{
© 2024 IEEE.  Personal use of this material is permitted.  Permission from IEEE must be obtained for all other uses, in any current or future media, including reprinting/republishing this material for advertising or promotional purposes, creating new collective works, for resale or redistribution to servers or lists, or reuse of any copyrighted component of this work in other works.
\par$^{1}$TUM eAviation Group, Department of Aerospace and Geodesy
\par $^{2}$Aerial Robotics Laboratory, Imperial College London
\par $^{3}$Laboratory of Sustainability Robotics, EMPA, Dübendorf, Switzerland}
\author{Luca Romanello$^{1,2}$, Daniel Joseph Amir$^{1}$, Heinrich Stengel$^{1}$,  Mirko Kovac$^{2,3}$, Sophie F. Armanini$^{1}$ \\}
}

\maketitle

\begin{abstract}
Underwater soft grippers exhibit potential for applications such as monitoring, research, and object retrieval. However, existing underwater gripping techniques frequently cause disturbances to ecosystems. In response to this challenge, we present a novel underwater gripping framework comprising a lightweight gripper affixed to a custom submarine pod deployable via drone. This approach minimizes water disturbance and enables efficient navigation to target areas, enhancing overall mission effectiveness.
The pod allows for underwater motion and is characterized by four degrees of freedom. It is provided with a custom buoyancy system, two water pumps for differential thrust and two for pitching. The system allows for buoyancy adjustments up to a depth of 6 meters, as well as motion in the plane.
The 3-fingered gripper is manufactured out of silicone and was successfully tested on objects with different shapes and sizes, demonstrating a maximum pulling force of up to 8 N when underwater. The reliability of the submarine pod was tested in a water tank by tracking its attitude and energy consumption during grasping maneuvers. 
The system also accomplished a successful mission in a lake, where it was deployed on a hexacopter. Overall, the integration of this system expands the operational capabilities of underwater grasping, makes grasping missions more efficient and easy to automate, as well as causing less disturbance to the water ecosystem. 

\end{abstract}


\section{Introduction \& Motivation}
\begin{figure}[t!]
    \centering
	\includegraphics[width=\columnwidth]{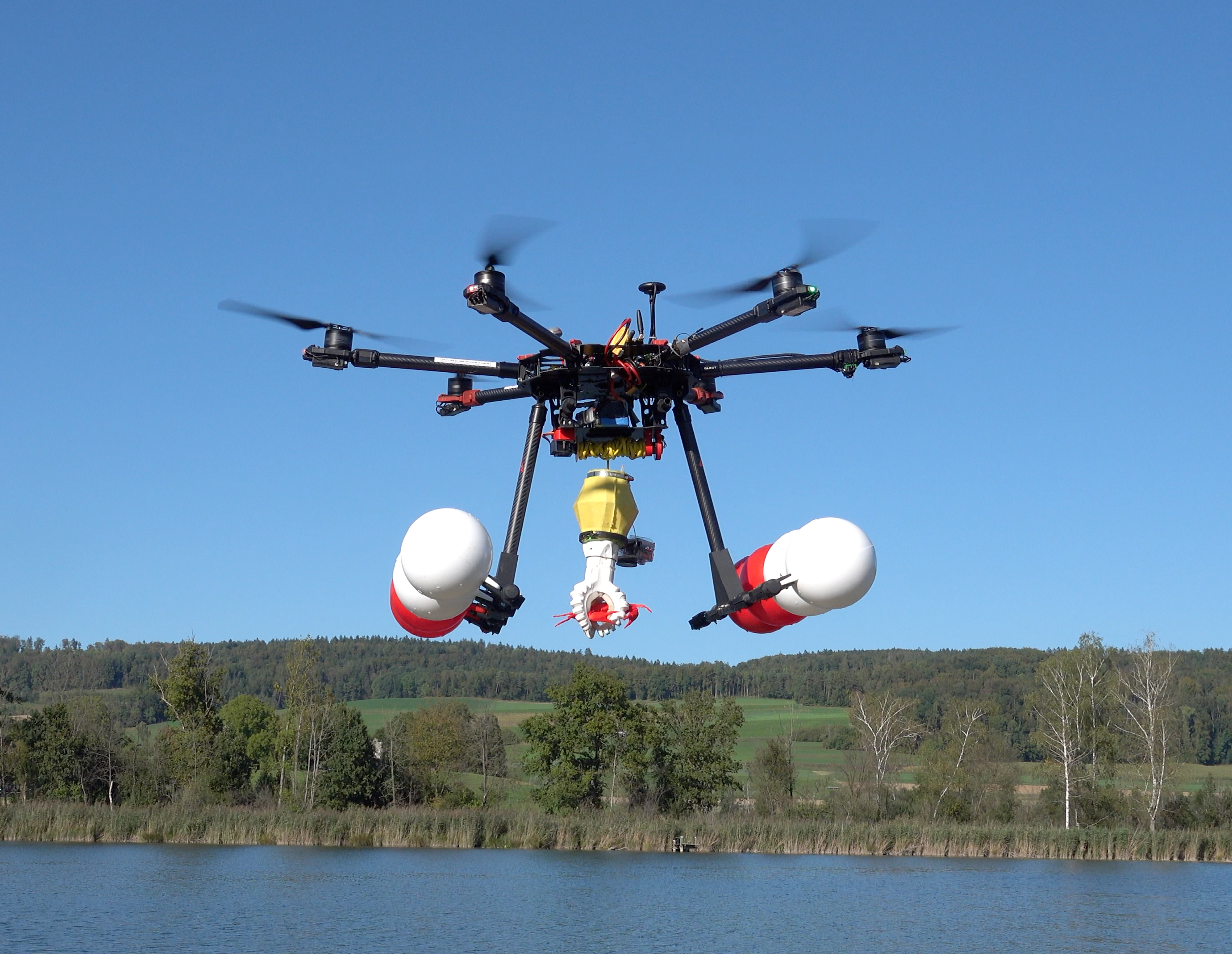}
    \caption{Outdoor tests on a lake in Switzerland. The system composed of the pod (yellow) and the gripper (white) is grasping a 3D-printed-crab while suspended on a hexacopter.}
\end{figure}
Utilizing subaquatic manipulators to grasp species and specimens in marine ecosystems in shallow water and on the seabed in lieu of submerged divers can 
minimize or avoid risks to divers, as well as enhance the automation of such underwater operations \cite{6907847,doi:10.1177/0278364920917203,ming2020grasping}.
Underwater grippers play a pivotal role in diverse underwater intervention scenarios. Their functions encompass grasping objects, whether they are floating on the water surface, submerged underwater, or resting on the seabed. They are also used for collecting biological research samples, recovering items like debris, and performing maintenance tasks in underwater industrial facilities ~\cite{galloway2016soft}.

It is worth noting that while the common underwater manipulators deployed for these applications display agility and flexibility, they lack stand-alone underwater locomotion capabilities. They are commonly affixed to remotely operated vehicles (ROVs) to reach their intended target.  

ROVs can be categorized into two main subgroups: underwater vehicle–manipulator systems (UVMS) and benthic crawlers \cite{9216533}. 
UVMS \cite{6907203,doi:10.1177/0278364920917203, ming2020grasping, ultragentle} are robots designed to manipulate underwater floating objects and have demonstrated successful grasping operations, even in deep-sea applications \cite{galloway2016soft,doi:10.1089/soro.2017.0097}. Benthic crawlers \cite{9216533}, by contrast, move along the seabed using tracks or wheels.

The utilization of large underwater systems can potentially disrupt the surrounding area of interest. Moreover, these systems are usually transported through the target area using surface vessels, requiring human crews for piloting, thus diminishing the feasibility of automation for these tasks.

We present a novel framework for agile underwater grasping tasks, merging the benefits of powered underwater mobility, as seen in ROVs, and the grasping capability of manipulators. Our approach consists of a compact submarine pod equipped with a soft 3-fingered gripper.
Instead of being carried across long distances by a water surface vessel or a ROV, the manipulator is deployed aerially to reach the target area.

The weight and size of earlier mentioned remotely operated vehicles pose transportation and handling challenges during missions. Deep-sea UVMs can weigh more than 100 kg and reach 1.5 m in length ~\cite{8027199,8593604,casalino,galloway2016soft}. Instead, if we consider even the most compact manipulators, they have a substantial weight, e.g. the manipulator described in Shen et al. \cite{doi:10.1089/soro.2019.0087}, weighs 2.5 kg, without considering the deploying vehicle. This weight and size are not compatible with the payload capacity of a standard multirotor.

Thanks to its low weight and dimensions, the pod-gripper system developed in this work can be mounted on a drone for rapid and less invasive underwater deployment.
Also, it excels at maneuvering into confined spaces and causing minimal disturbance to the surrounding marine life during missions due to its small size and soft structure.

The gripper-equipped pod is attached to the deployment drone through a wired cable and a winding system. Once the location of a possible ecological mission involving grasping maneuvers is defined, the multirotor will fly to the location of interest and land on the water. Upon landing, the drone is switched off, the winding system activated and the gripper lowered into the water, to start its mission. The submarine pod can now be controlled remotely via FPV (first-person view) to reach its target location and perform its grasping mission using the gripper. 

Once the underwater mission has been completed, the pod is retrieved and the drone can either fly back to land or continue to reach another mission location, either within the same water body or in another one. 
To achieve rapid deployment, flight capability is an evident advantage, while at the same time allowing access to remote water bodies where the shoreline is not accessible. This operational concept is based on our previous work \cite{debruyn2020medusa}, where it was developed for a different type of environmental application.

\begin{figure*}[t]
    \centering
	\includegraphics[width=2\columnwidth]{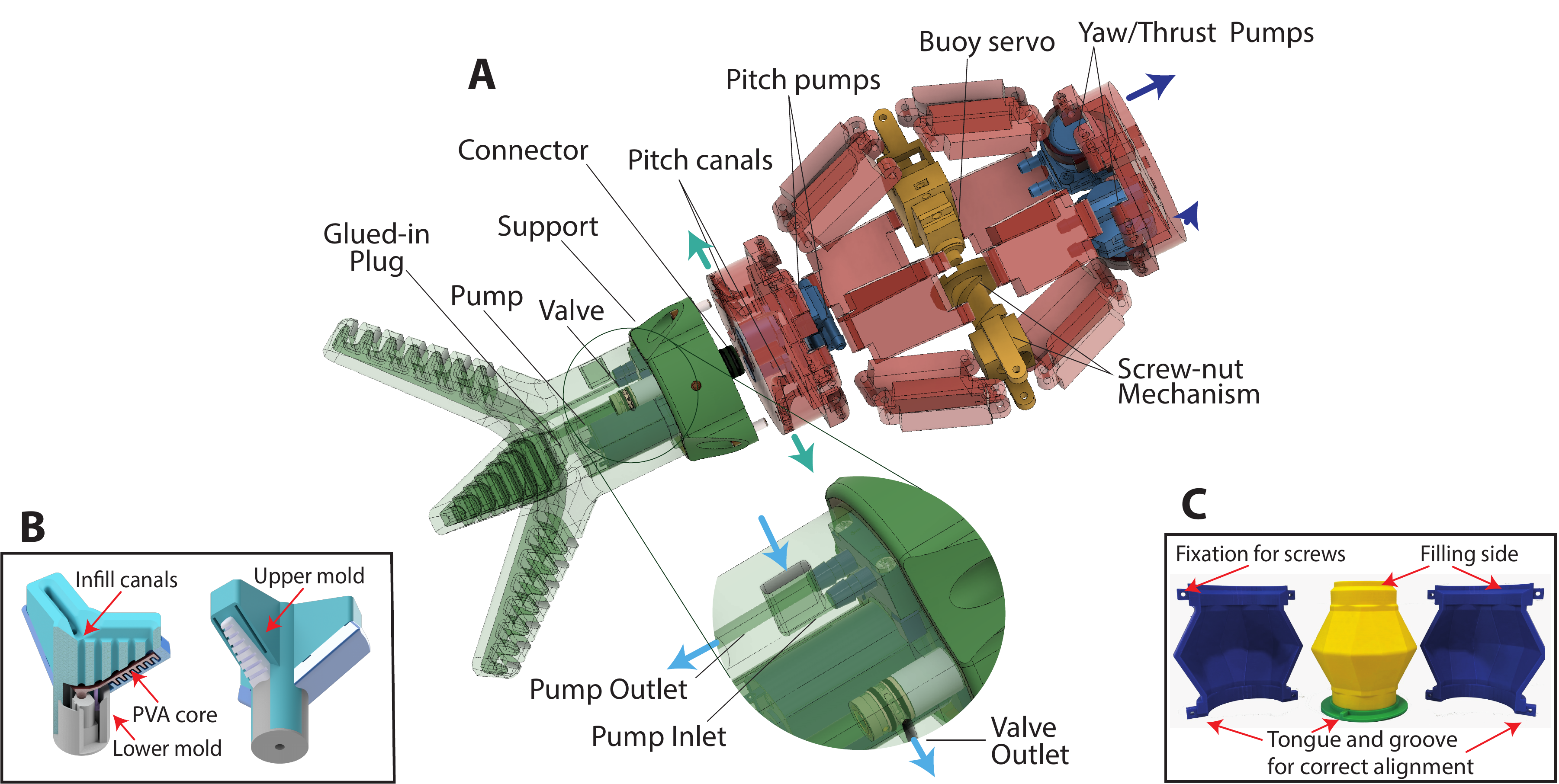}
    \caption{A) Pod-gripper system design. B) Casting mold for the gripper and C) pod. For
the case of the pod, outer-shell (blue) inner-shell (red) and base-plate (green).}\label{pod_design}
\end{figure*}

Utilizing UAVs for underwater missions enhances efficiency, compactness, and effectiveness. Our approach minimizes ecosystem disruption and boosts energy efficiency compared to traditional ROV methods. Once the aerial vehicle has landed on the water surface and is powered down, the deployment of the pod-gripper mechanism and the grasping operation itself cause only minimal disturbance, as well as requiring only minimal energy for underwater locomotion.




 Our main contributions include:
 \begin{itemize}
    \item The development of a novel framework for underwater grasping operations which has the potential to minimize disturbances to marine ecosystems.
    \item The design, development and testing of a novel underwater submarine pod equipped with an active buoyancy control mechanism.
    \item The development and testing of a compact and lightweight 3-finger soft underwater gripper, designed for safe underwater grasping operations.
    \item Outdoor testing of the full system with deployment from a multirotor platform.
 \end{itemize}
 
\section{Concept \& Design}

The proposed framework aims to employ a gripper for grasping operations for rapid deployment via an unmanned aerial vehicle (UAV).
For this system to be effective in addressing the grasping operation, it necessitates the ability to execute underwater movement, including fine adjustments for precise target approach. These two core functionalities, grasping and locomotion, have been independently developed. Grasping is achieved through a multi-fingered gripper, while underwater locomotion is facilitated by a submarine pod equipped with control capabilities for essential locomotion parameters: water depth, differential thrust, and pitch.
The pod-gripper system has been designed as a standalone unit, independent of the host platform. This modular design approach allows for versatility and ease of integration with various host platforms and simplifies maintenance and upgrades when needed. 

\subsection{Design Requirements} \label{subsect:concept}

In view of the aerial deployment of the of the system, the design of the pod-gripper system follows the paradigm for designing aerial-aquatic vehicles~\cite{zufferey2022between}:
\begin{description}
   \item[Safe] The design must take into account material compatibility with grasping fragile samples, emphasizing a shape and structure that prevents damage.
   \item[Light-weight] Due to their integration on submersible rovers, underwater grippers are ordinarily designed without weight constraints. However, for aerial deployment, weight becomes a crucial factor as it influences the efficiency of the mission. Thus, the goal is to minimize of the weight of the gripper-pod system.
   \item[Compact] The gripper-pod must maintain neutral buoyancy for effective underwater locomotion and grasping. Consequently, the design must prioritize compactness, ensuring that the system occupies the smallest possible volume.
   \item[Platform-independent] The system necessitates a standalone design with seamless compatibility for integration with various UAV platforms. It should incorporate integrated communication with the external environment. 
   \item[Modular] Both pod and gripper need to be design modular, allowing easy access for adjustments, upgrades, and maintenance, including hardware inspections and algorithm updates on the micro-controller.

\end{description}

This design approach enhances reliability and facilitates the integration of the system with different drones or the use of various ecological sensing units on the same drone, as demonstrated in \cite{farinha2022off}.

\subsection{Design of the Gripper}

In order to grasp biological specimens successfully, several methods can be applied. A soft robotic design allows the gripper to conform around the object being grasped \cite{laschi2017soft} and ensures gentle handling of delicate specimens, substantially reducing the risk of damage to the biological samples. This is achieved by creating an enveloping grasp, which can increase the robustness of the grasping process \cite{dollar2005towards} although the exerted forces are relatively small.
Out of the great variety of soft robotic technologies in existence today, we opted for fluidic elastomer actuators (FEAs), which allow us to grasp objects through direct actuation using pressurized fluids \cite{shintake2018soft}.

The three-fingered gripper is a homogeneous silicone rubber structure designed for grasping objects with primary dimensions ranging from approximately 40 mm to 90 mm. The use of silicone rubber is advantageous due to its exceptional bio-compatibility, resulting in minimal environmental impact on the water ecosystem \cite{shit2013review}. Opting for three fingers allows us to maintain a lightweight design while achieving great precision, as evidenced in Feix et al. \cite{FingersPrecision}. This configuration ensures an effective grasp by securely encasing the object being held. In terms of specifications, each finger comprises seven chambers, each measuring 3.5 mm in length, resulting in an overall finger length of approximately 70 mm. This design characteristic enables the gripper to safely handle objects with a diameter of up to 90 mm.



The described FEA-fingers are all part of the same continuous silicone rubber structure. Although a design with individual detachable fingers could allow for greater adaptability to different grasping tasks \cite{9216533,doi:10.1089/soro.2022.0215,phillips2018dexterous,10122042}, our choice was to pursue a single-part design. This decision was made to facilitate the creation of a lightweight and compact gripping module.
Furthermore, this choice facilitated the integration of onboard electro-pnuematic systems being housed by the silicone rubber structure as well. This way, the gripper does not require external pressure sources, tubing, and a valve system, unlike most existing grippers. The resulting lightness and compactness of the overall gripping system make it particularly suited for aerial deployment.
The pump inside the gripper is attached to the channels via a press fit and held in place by such, seen in \cref{pod_design}A. Furthermore, a valve is placed in an outlet channel of the silicone rubber structure, allowing for storage and release of the pressurized water actuating the gripper independently of the pump's actuation.
An end cover with a connector system seals the compartment housing the pump, while also allowing for the electronic cables to pass into the pod (seen in \cref{pod_design}A). 

A plug seen in \cref{pod_design}A is glued in a leak of the FEA channeling existing due to the molding manufacturing. 

\subsection{Design of the Pod}

While numerous different designs have been proposed for underwater pods and autonomous underwater vehicles (AUVs), most of the systems developed so far are either too heavy to be carried by a drone \cite{ming2020grasping} or do not have the capability to grasp objects underwater \cite{debruyn2020medusa} \cite{plum2020sauv} \cite{katzschmann2018robofish}. \\
Our objective is to obtain a lightweight and small system, which will have a minimum impact on the surrounding environment when deployed. Therefore we propose a vehicle with a fully soft flexible cover, which reduces the risk of damaging the ecological environment under investigation during missions. The flexible skin, which weighs about 110 g, also allows for a novel buoyancy control mechanism, as discussed in \cref{sucsec:skin}.

The four degrees of freedom the pod has been designed with allow the gripper to nimbly move around the area of interest. The buoyancy control is done via a screw-nut mechanism, driven by a servo (marked in yellow in \cref{pod_design}A). By actuating this mechanism, the shape of the skin can be changed and consequently the volume and buoyancy of the entire structure.\\
Because of the simple actuation and small number of moving parts compared to other mechanisms, an umbrella structure with the shape of a hexagonal bipyramid ~\cite{stoeffler2023jet} was chosen, as shown in \cref{pod_design}A.
The volume of the pod is changed by deforming a soft cover, in the vein of Hritwick et al. \cite{hritwick2018orumbot}, where expansion and contraction of a soft cover is used for underwater thrust generation. Given that in our design the volume is changed for buoyancy control, a higher pressure acts on the soft cover. Therefore the silicone needs to be more supported by the rigid structure beneath it, so that the shape of the soft cover can be sustained. 

\begin{figure}[htbp]
    \centering
	\includegraphics[width=\columnwidth]{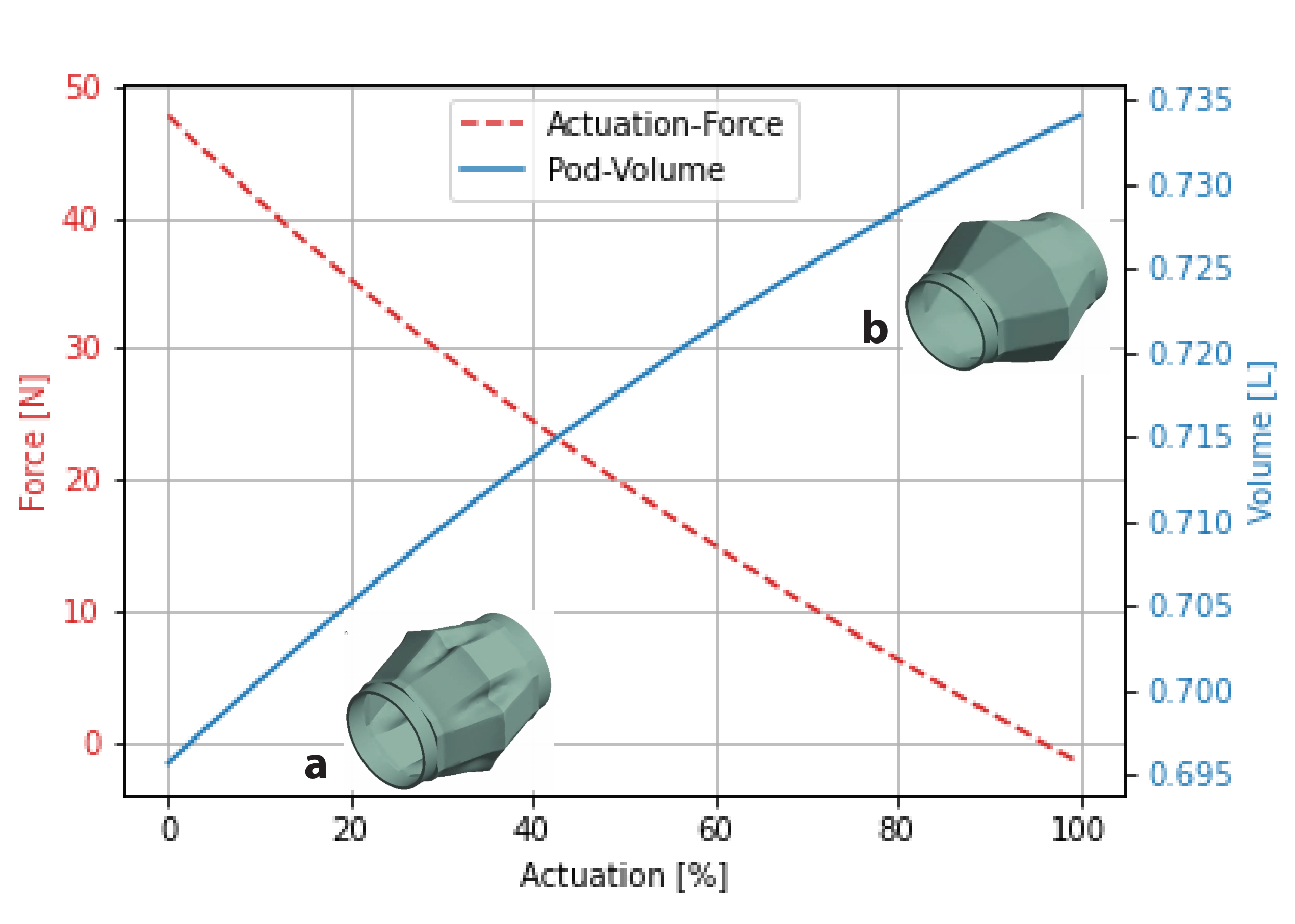}
    \caption{Calculated actuation-Force and Volume of the pod while actuating the buoy servo system at a constant water depth of 0.5m.} \label{fig:volume_no_water}
\end{figure}

The volume changes with respect to the actuation as shown in \cref{fig:volume_no_water}. In the initial design process, the deformation of the silicone skin under water pressure was neglected and only geometric relations were considered.
To estimate the course of the actuation force, the force caused by the pressure acting on the undeformed surface was calculated, based on the leverage arm of the mechanism.

Upon full servo actuation, the silicone skin initially occupies its maximum volume \cref{{fig:volume_no_water}}b. When the servo rotates, the silicone skin stretches along the pod's length, causing the skin between the support arms to fold inward until it reaches its maximum extent \cref{{fig:volume_no_water}}a. This folding action, akin to an umbrella, reduces the occupied volume. This results in a volume change of 5.7\% when considering the pod alone and 3.6\% when accounting for the entire system.

For the forward motion of the pod underwater, two centrifugal water pumps are used, which are mounted at the back inside the pod, as seen in \cref{pod_design}A. To achieve a maximum impulse, the pumps suck water from the front of the pod and eject it backwards in the opposite direction. The pumps can be controlled separately to achieve differential thrust, thus generating a yawing moment.\\
To control the pitch of the pod, two smaller pumps are used, which are mounted at the front of the pod. Also these pumps suck and eject water in opposite directions. Tubes and 3d-printed channels inside the parts are used to guide the water in the intended direction. 

\subsection{Electronics}

\begin{figure}[htbp]
    \centering
	\includegraphics[width=0.8\columnwidth]{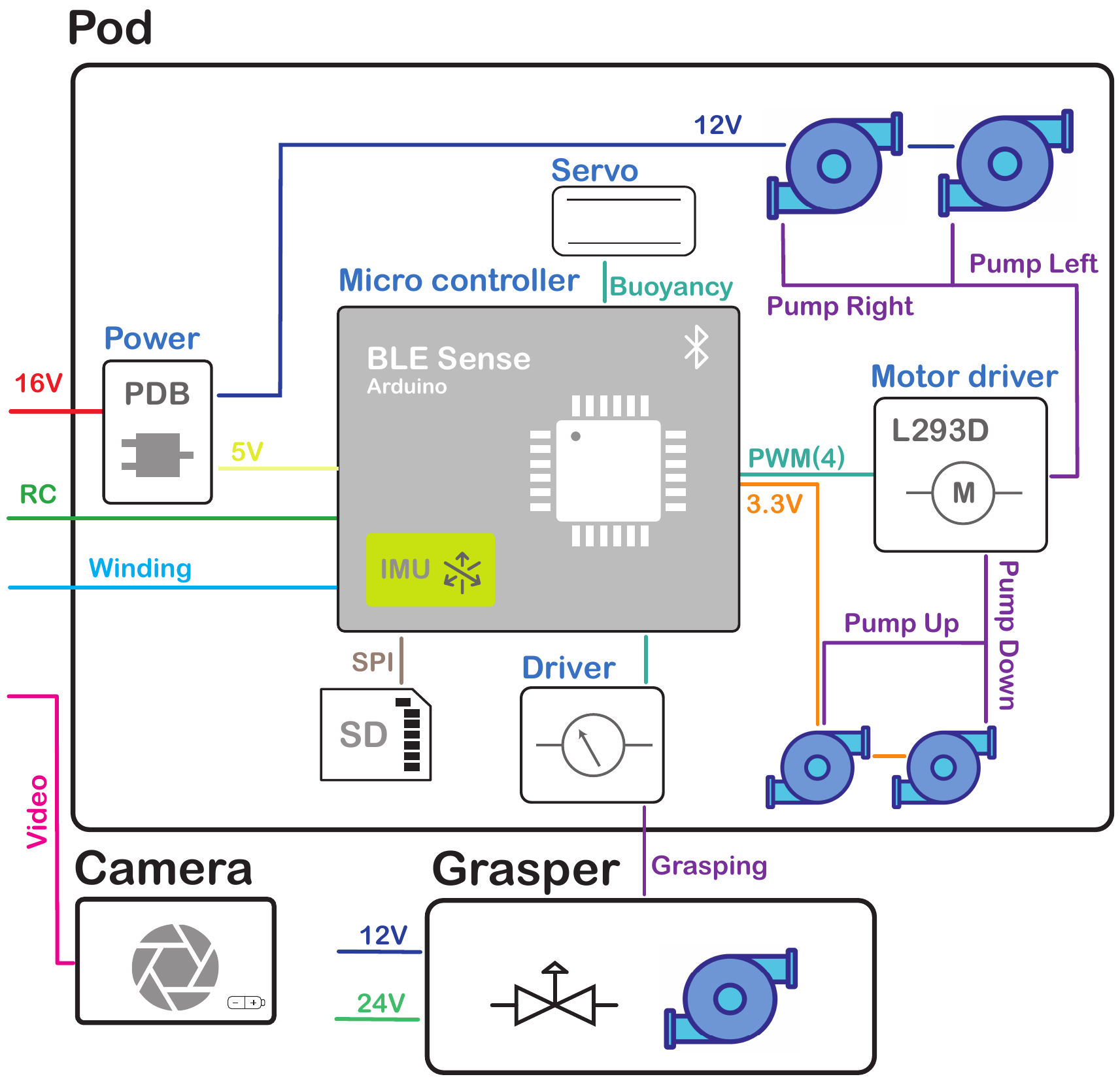}
    \caption{Electronics' scheme.}
\end{figure}

As introduced earlier, the system, including pod and gripper, uses the tether to communicate with the multi-rotor platform, which provides energy and RC communication, and receives the FPV camera video signal and the winding PWM output.
The Pod electronics package includes a power distribution board to power the Arduino board with 5V and 12V for the pod's internal control pumps. The microcontroller is equipped with an integrated IMU, enabling the tracking of system attitude. It interfaces with six RC channels, which represent thrust, yaw, pitch, buoyancy, cable winding, and grasping inputs. Using an open-loop system, it transforms these input signals into five PWM output signals. Two PWM signals control thrust and yaw for the two larger pumps, one PWM regulates pitch for the two smaller pumps, and the remaining three signals govern the buoyancy, the grasping and the cable winding subsystems.

The micro-controller is soldered onto a custom PCB contained in the pod. The board contains all the Arduino board's connections, including the two motor drivers for the big and small pumps, and the driver mosfet for controlling the gripper pump and valve. 
 
\section{Manufacturing \& analysis} 

\subsection{Manufacturing}

The covering skin of the gripper and pod was manufactured out of silicone rubber, which allowed for easy manufacturing using molds \cite{marchese2014recipe}. 
A high-curing-time silicone rubber with soft range hardness (shore 30A) was selected as a material, based on similar work in the literature~\cite{pagoli2021review}.
In particular, the use of the chosen material facilitated the attainment of an optimal rubber quality characterized by minimal air entrapment and complete infill of the intended geometric structure. 


The gripper cores for the casting process were 3D printed out of polyvinyl alcohol (PVA) which, thanks to its water solubility, is suitable for creating the internal structures needed for the FEA cavities. 

The casting molds were likewise 3D-printed out of PLA. A multi-segmented design allowed for easy disassembly and reuse of the whole casting mold, avoiding undercuts within the structure \cref{pod_design}A. 

To manufacture the pod, a soft outer skin and a rigid PLA-based skeleton structure were employed. The same silicone was used for both the soft outer skin and the gripper, following a casting method utilized in similar studies in the literature \cite{stoeffler2023jet}. A four-part mold was constructed for casting the silicone, as shown in \cref{pod_design}C. The outer mold shell consists of two parts for easier demolding, while the inner shell remains a single piece thanks to silicone's flexibility.

\subsection{Analysis of the Grasping Mechanism}

\begin{figure}[!h]%
\captionsetup{width=\linewidth}
    \centering
    \includegraphics[width=0.9\columnwidth]{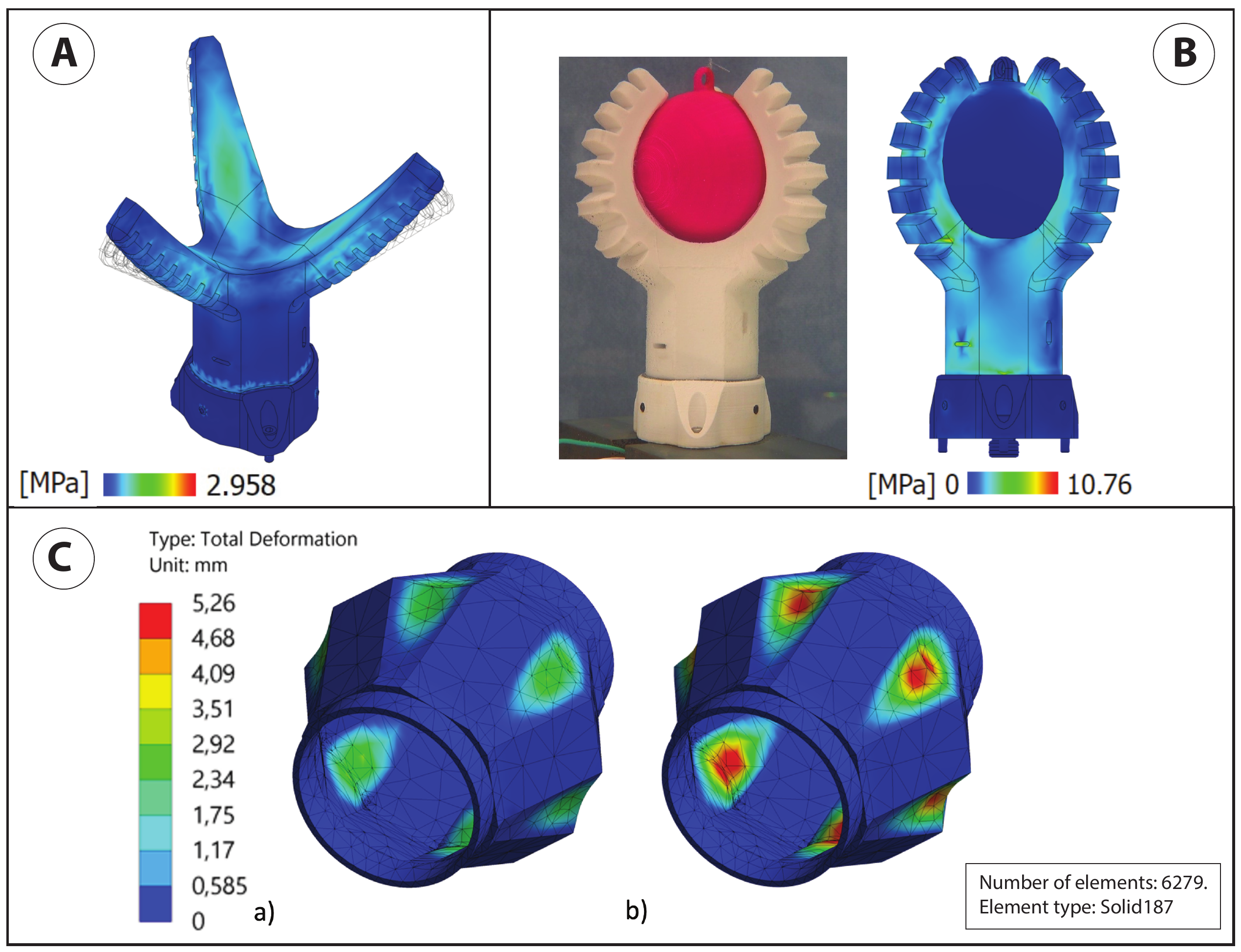}
   \caption{ A) Von Mises stress analysis due to 0,1 MPa of pressure. B) Gripper grasping a sphere underwater while the latter is pulled upwards. Stress analysis according to von Mises method. C) Deformation of the soft skin at a depth of, respectively, 0.5m (a) and 1.5m (b). Number of elements: 6279. Element type: Solid187.}%
   \label{fig:grasps_stress}%
\end{figure}%

Von Mises stress analysis was performed to assess the gripper's dynamics. One noteworthy observation in both simulations and real-world testing is the occurrence of buckling at the central regions of the plane sections, as indicated by the elevated von Mises stress levels (green areas) in \cref{fig:grasps_stress}A. This discovery implies that the gripper's overall geometry could be enhanced by reinforcing the areas where buckling is most pronounced, such as by adopting a more cylindrical shape for the fingers.



Finite element simulations were conducted using ANSYS to model the scenario in which the gripper is holding or retaining a sphere while a 10 N pulling force is applied to the sphere. In these simulations, the gripper was considered fixed at the end-cover, and a pressure of 0.15 MPa was applied within the cavities.
The obtained von Mises stress is shown in \cref{fig:grasps_stress}B. As expected, a concentration of stress can be seen at the contact surface between the fingers and the grasped objects. Aside from these peak stresses, the stress distribution is rather homogeneous, with no areas of stirringly excessive strain.

The gripper in this project took approximately 20 seconds to fully close and 25 seconds to open, as seen in the supplementary video.


\subsection{Analysis of the silicone skin} \label{sucsec:skin}
The pod's depth can be changed via buoyancy control, which is achieved by changing the volume of the pod, and therefore also the shape of the outer skin. Because of the highly elastic behavior of the silicone, the buoyancy control process is heavily influenced by the water pressure acting on the pod and the depth of the pod inside the water.
\begin{figure}[htbp]
    \centering
	\includegraphics[width=\columnwidth]{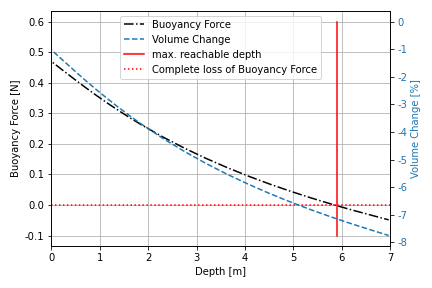}
    \caption{Maximal reachable depth of the pod due to the deformation of the skin, obtained from the simulation results.}
    \label{depth}
\end{figure}

To gain a better understanding of this behavior, the silicone skin was investigated by means of a finite element method simulation in Ansys. Hereby, the shape of the skin at the maximum pod's volume was examined as shown in \cref{fig:grasps_stress}C.
The load-case considered in the simulation consists of a force boundary condition, i.e. applying a pressure on the whole outer surface of the skin.
The displacement at the two circular ends, where the skin is clamped to the rigid body, is set to zero. Where the skin is supported by the rigid arms beneath, no deformation normal to the surface is allowed.
The deformation of pod's soft skin was simulated in a pressure range from 0 to 0.07 MPa, which corresponds to a water depth from 0 to 7 meters.
In \cref{fig:grasps_stress}C, the deformation of the pod at 0.5 m and 1.5 m depth is shown.\\
From the displacement of the skin under different pressures, the maximum volume of the pod under different pressures was obtained.
Each applied pressure, and the corresponding change in volume, can be mapped to a certain depth. From this mapping we obtain the change in maximum volume over depth. When we compare the maximum volume of the pod at a particular depth with the initial volume at neutral buoyancy, we can calculate the buoyancy force acting on the pod. The results can be seen in \cref{depth}.

As shown in the plot, the volume of the pod decreases with depth inside the water, by up to almost 7\%. Therefore also the buoyancy force decreases, up to the point where the pod is no longer able to sustain a neutral buoyancy. This marks the maximum depth the pod can reach and return from, which corresponds to almost 6 meters.

\subsection{Mass distribution analysis} \label{chap:masses}

\begin{figure}[htbp]
    \centering
	\includegraphics[width=\columnwidth]{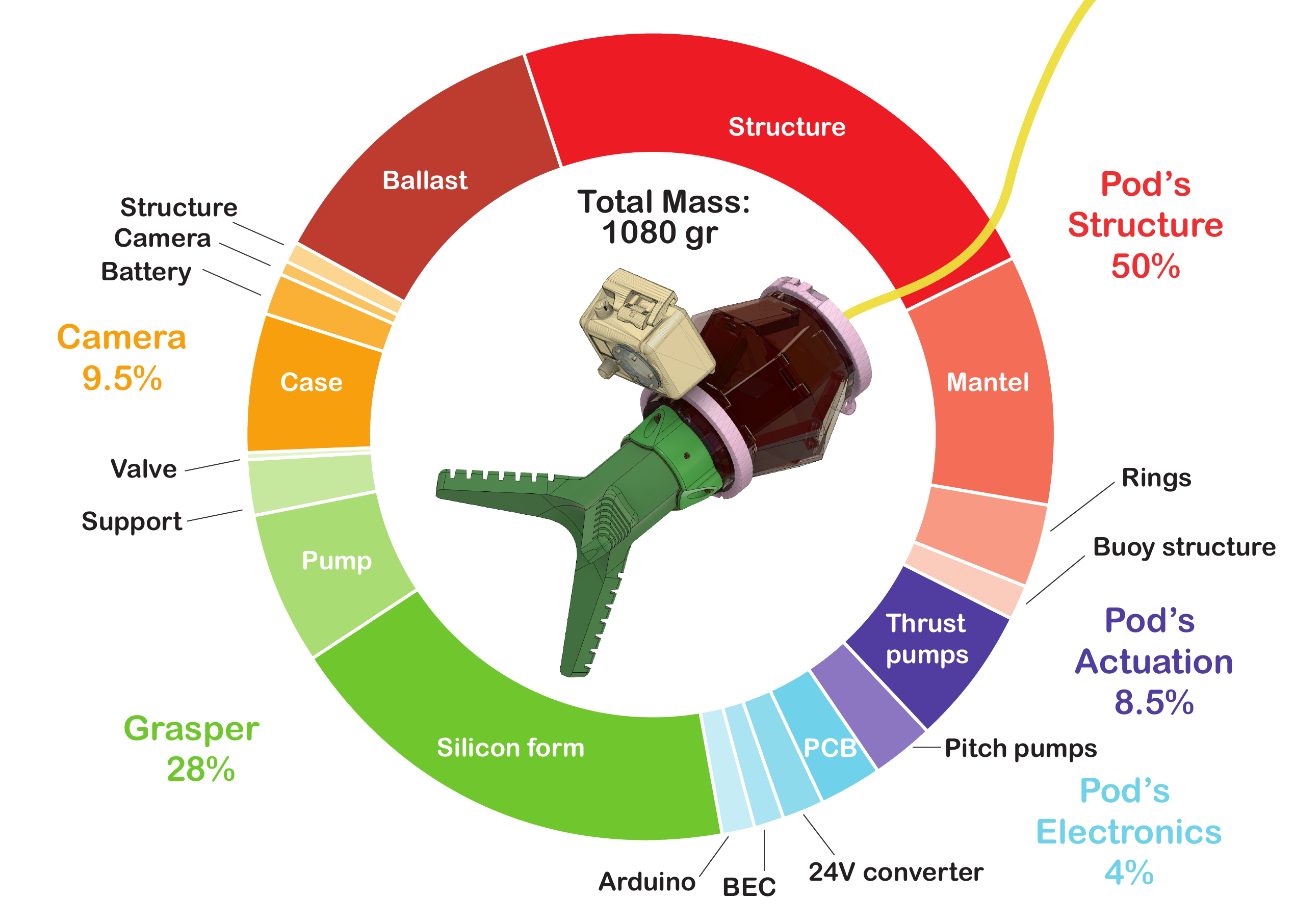}
    \caption{Distribution of mass of each element within the pod-gripper system.}
\end{figure}

As outlined in \cref{subsect:concept}, prioritizing lightweight design was a primary consideration. To facilitate underwater piloting, we affixed an FPV camera with a waterproof case to the pod. This component addresses vehicle dynamics, leveraging its mostly empty interior for underwater control. Given the gripper's negative buoyancy, designing the pod with positive buoyancy would lead to undesired robot vertical orientation, contrary to our objectives. The camera case, buoyant by design, offsets the gripper's negative buoyancy. Additionally, as the pod lacks roll actuation, the camera case serves as embodied feedback to maintain roll stability. However, the camera introduces drawbacks: it amplifies potential drag-induced forces opposing forward thrust and provides feedback on pitch, beneficial for horizontal orientation but diminishing pitching actuation effectiveness. 
The overall system mass is 1080 g. During testing, a cable for UAV connection and a custom cable winding system were added. Additionally, once the gripper is activated underwater, it retains water in its channels, increasing the overall mass by an amount equivalent to a quarter of the silicone form, i.e. 50 g. The total mass after a grasping operation would be approximately 1400 g. 

\section{Tests and results}\label{tests_gr}

Tests have been performed in a water tank as well as on a lake. Both the grasping capabilities of the gripper and the performance of the pod have been characterized to validate the reliability of the system.

\subsection{Gripper testing}
\label{chap:gripper_test}
\begin{figure}[!h]%
\captionsetup{width=\linewidth}
    \centering%
    \hspace{0.01\linewidth}
    \begin{subfigure}[t]{0.2\linewidth}
    \includegraphics[width=\linewidth]{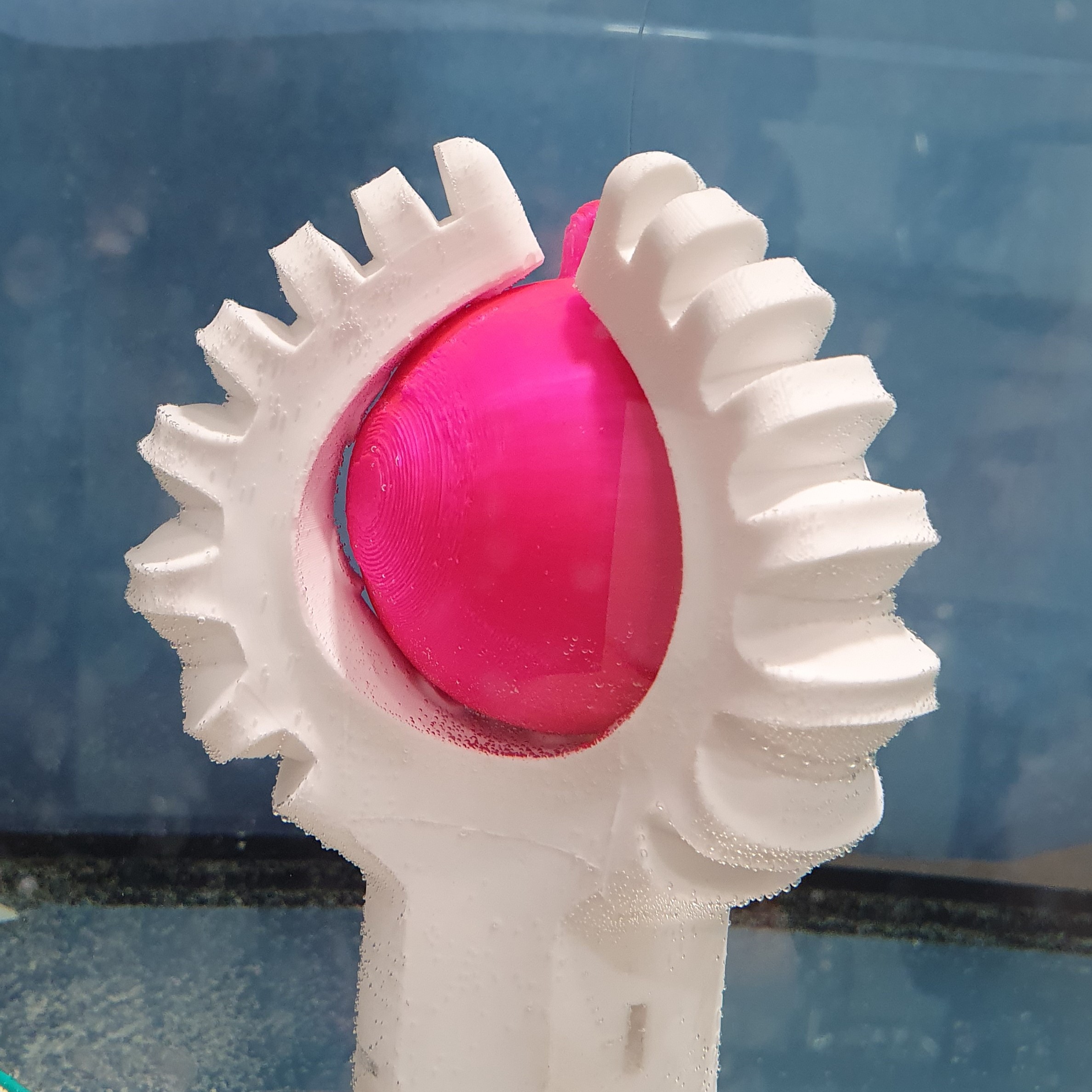}
    \end{subfigure}
    \hspace{-0.01\linewidth}
    \begin{subfigure}[t]{0.2\linewidth}
    \includegraphics[width=\linewidth]{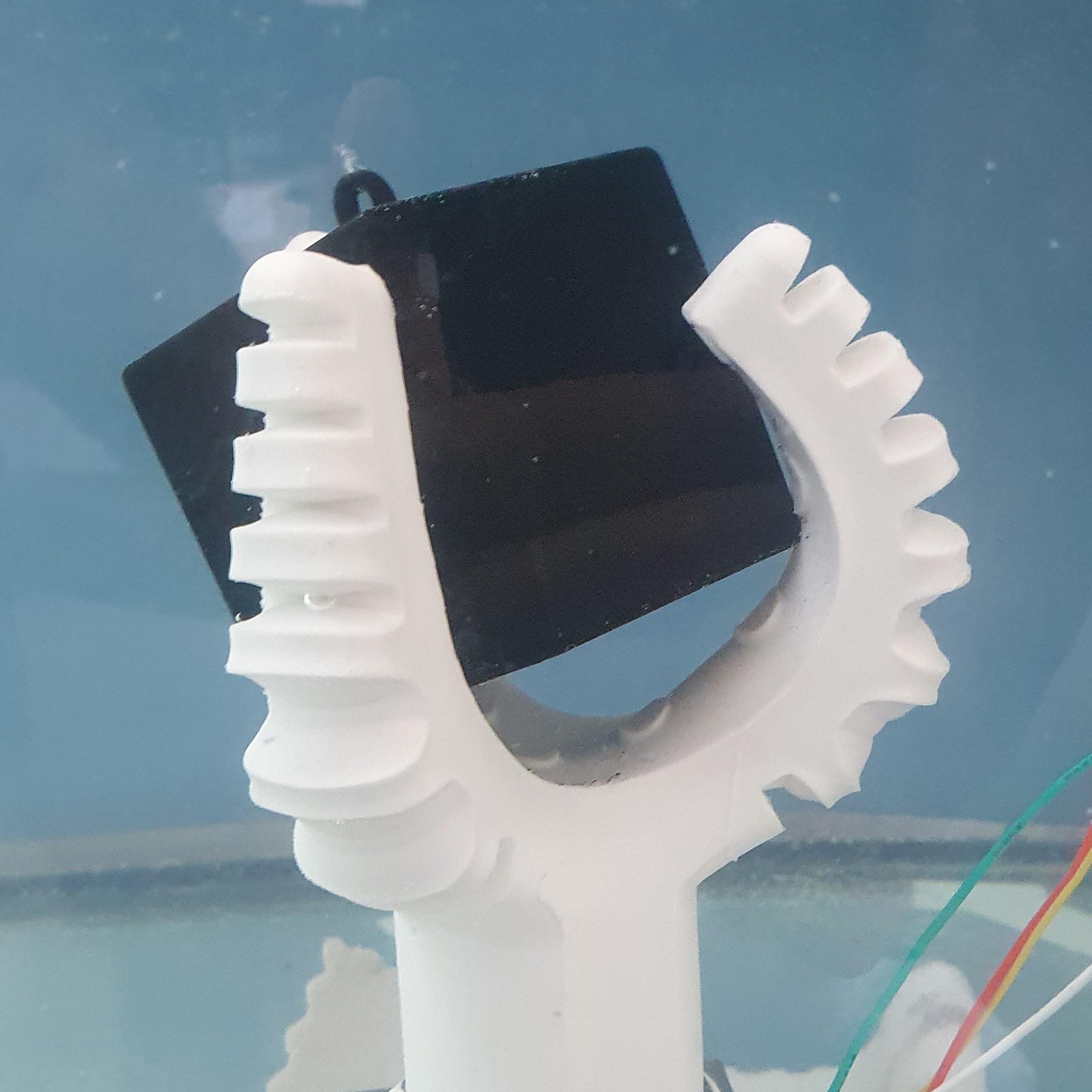}
    \end{subfigure}
    \hspace{-0.01\linewidth}
    \begin{subfigure}[t]{0.2\linewidth}
    \includegraphics[width=\linewidth]{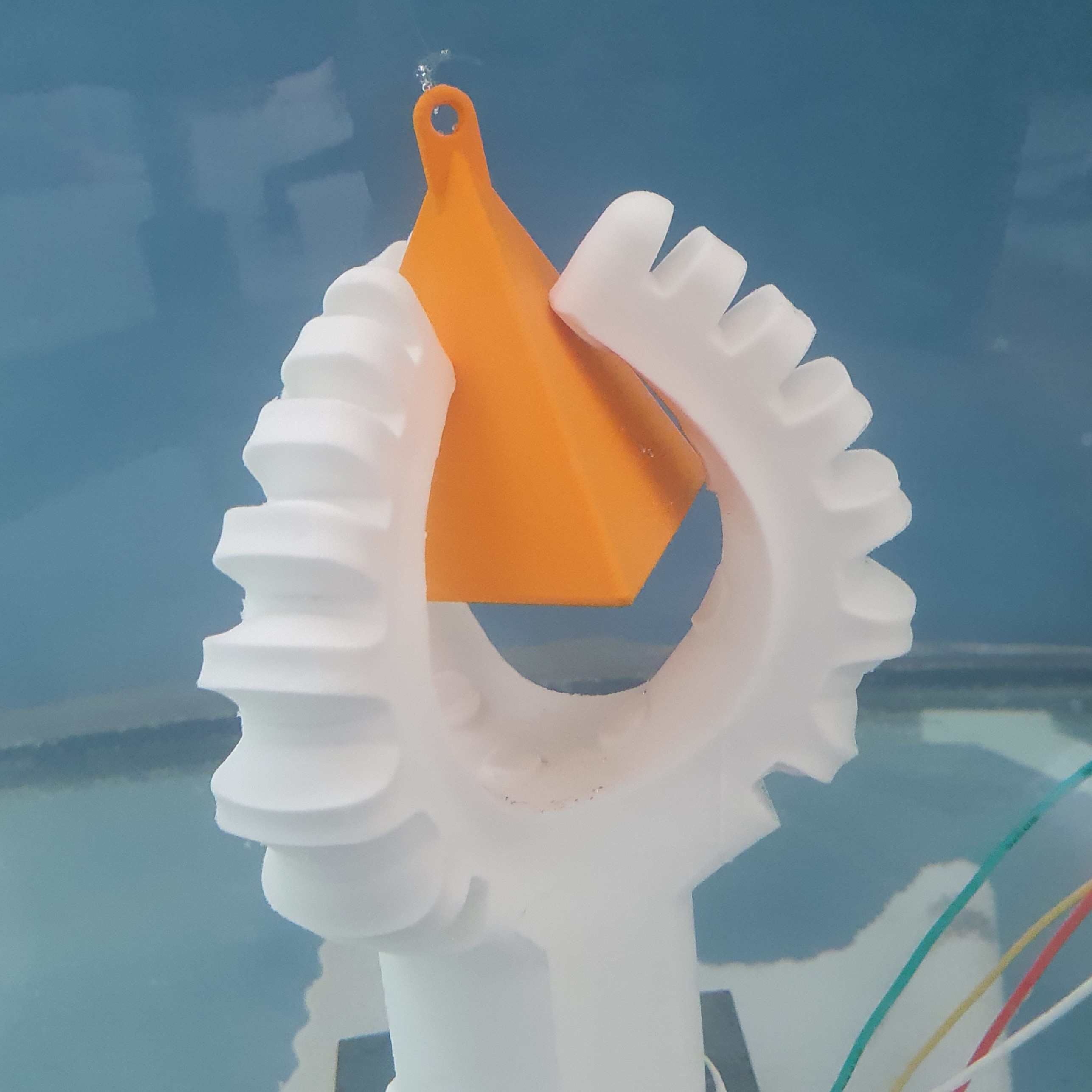}
    \end{subfigure}
    \hspace{-0.01\linewidth}
    \begin{subfigure}[t]{0.2\linewidth}
    \includegraphics[width=\linewidth]{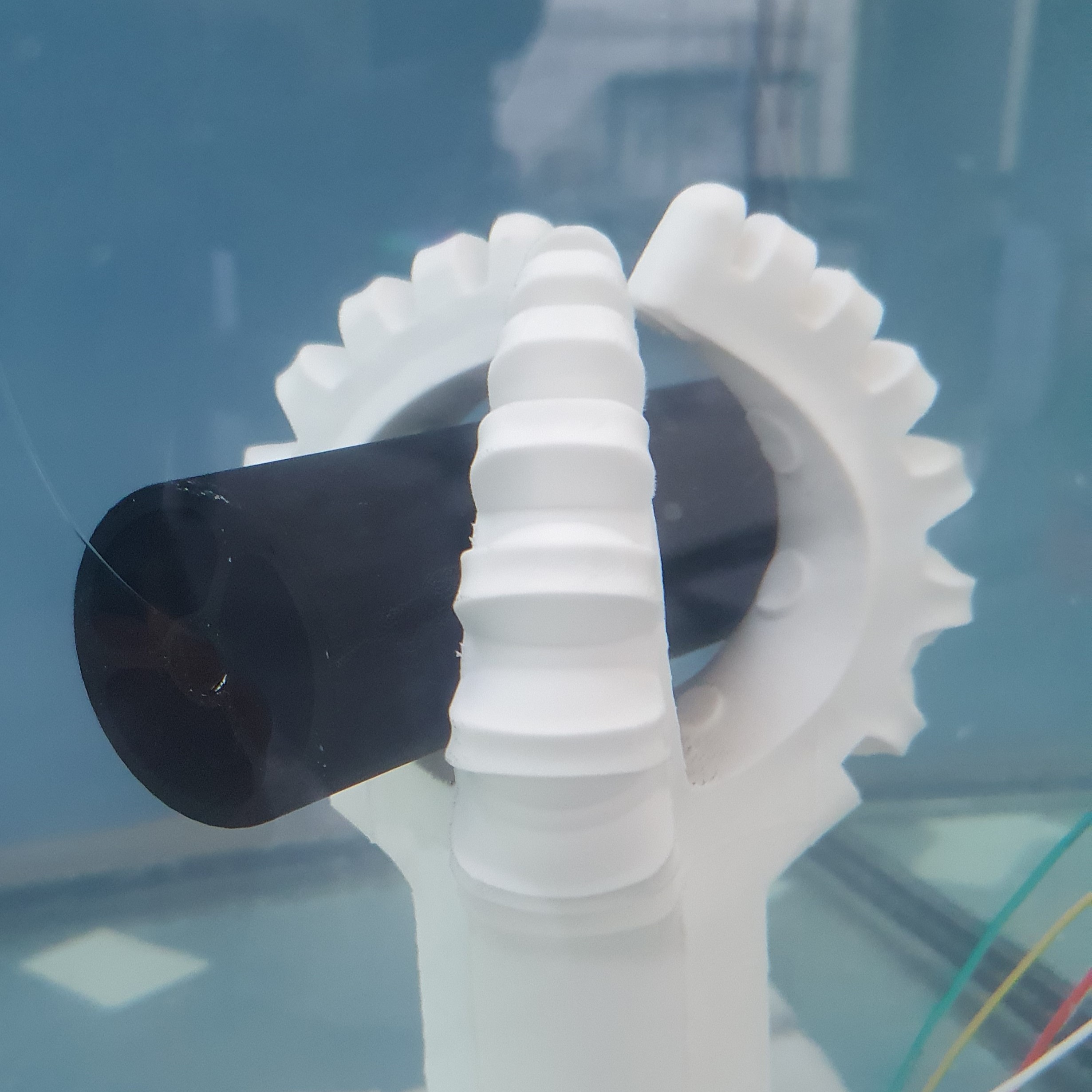}
    \end{subfigure}\\
    \vspace{2mm}
    \hspace{-4mm}
    \includegraphics[width=0.87\linewidth]{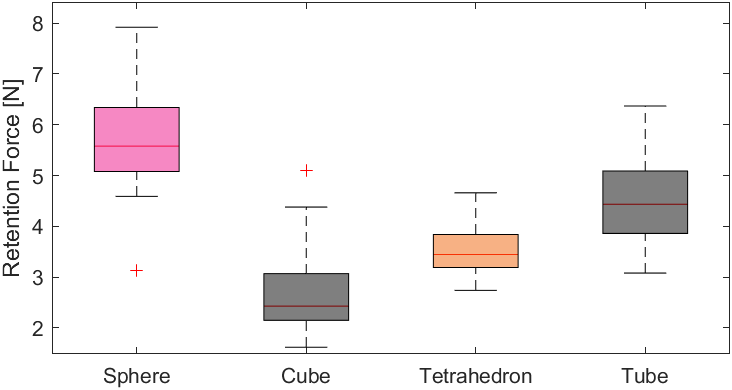}\\
    \includegraphics[width=0.9\linewidth]{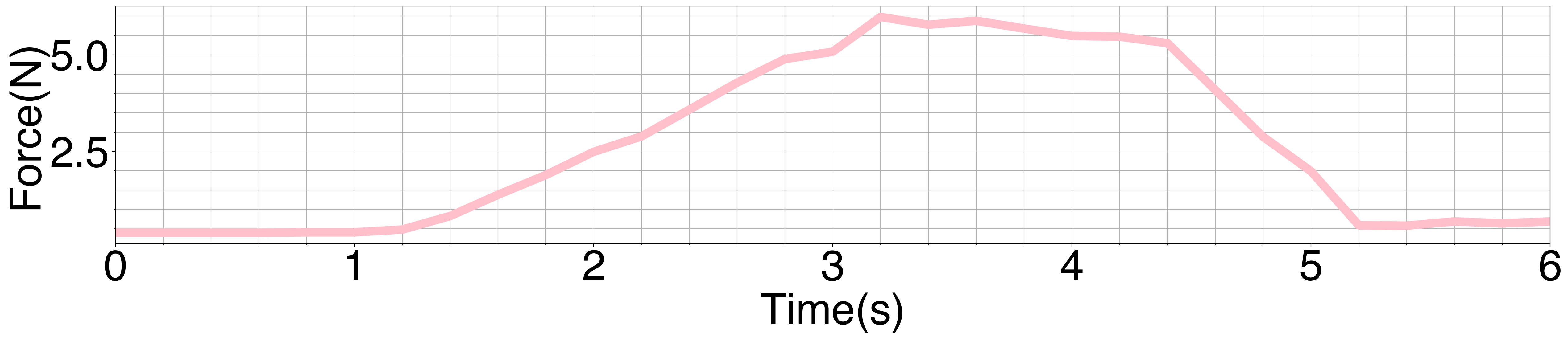}
    \caption{Pictures taken during the grasping experiment of sphere (60 mm diameter), cube (60 mm edge length), tetrahedron (60 mm edge length), and tube (130 mm long, 40 mm diameter) (top). Result of the pulling test with different geometries. The median, quartiles and outliers (red crosses) have been marked. (middle). Pulling force over time for the case of the sphere (bottom).}%
    \label{fig:geometries_plot}%
\end{figure}

In order to assess the gripper's functionality and performance, we carried out experiments to measure its retention force. This force represents the maximum strength at which the gripper can hold onto an object being pulled away.  
For these tests, the gripper was fixed underwater in a vertical position, and the grasped object was pulled upwards to remove it from the hold of the fingers \cref{fig:geometries_plot}. Weight and buoyancy were taken into account as shown in Equation 1: 
\begin{equation}
F_{Retention} = F_{Pulling,max} - F_{Weight} + F_{Buoyancy} 
\end{equation}
Several objects were tested, i.e. a sphere, cube, tetrahedron, and tube of similar dimensions. The results of this experiment can be seen in \cref{fig:geometries_plot}:middle.  The retention force varies noticeably, with the highest force reaching 8 N for the sphere and the lowest for the cube, which has a median value half that of the sphere.
For more detailed insight on the grasping behaviour, we also assessed the grasping behaviour over time. 
 Results are depicted at the bottom of \cref{fig:geometries_plot}, which shows the measured pulling force over time during the removal of a grasped sphere (60 mm) from the gripper, at a constant velocity. During the grasping process, strain and stress holding the object in place build up until reaching a plateau, where most of the grasping force transmission is perpendicular to the object's trajectory and friction seemingly dominates over form closure.


An important performance factor obtained from the testing phase is the ratio of object mass to gripper mass. The gripper weighs 300 g on land, whereas when submerged the whole unit has a buoyancy of 0.1 N. As shown in our experiments (see \ref{fig:geometries_plot}), the maximum weight the gripper can lift underwater is approximately 800 g. The resulting lifting ratio underwater is thus around 8. 

\subsection{Testing the Pod}
\begin{figure}[htbp]
    \centering
	\includegraphics[width=\columnwidth]{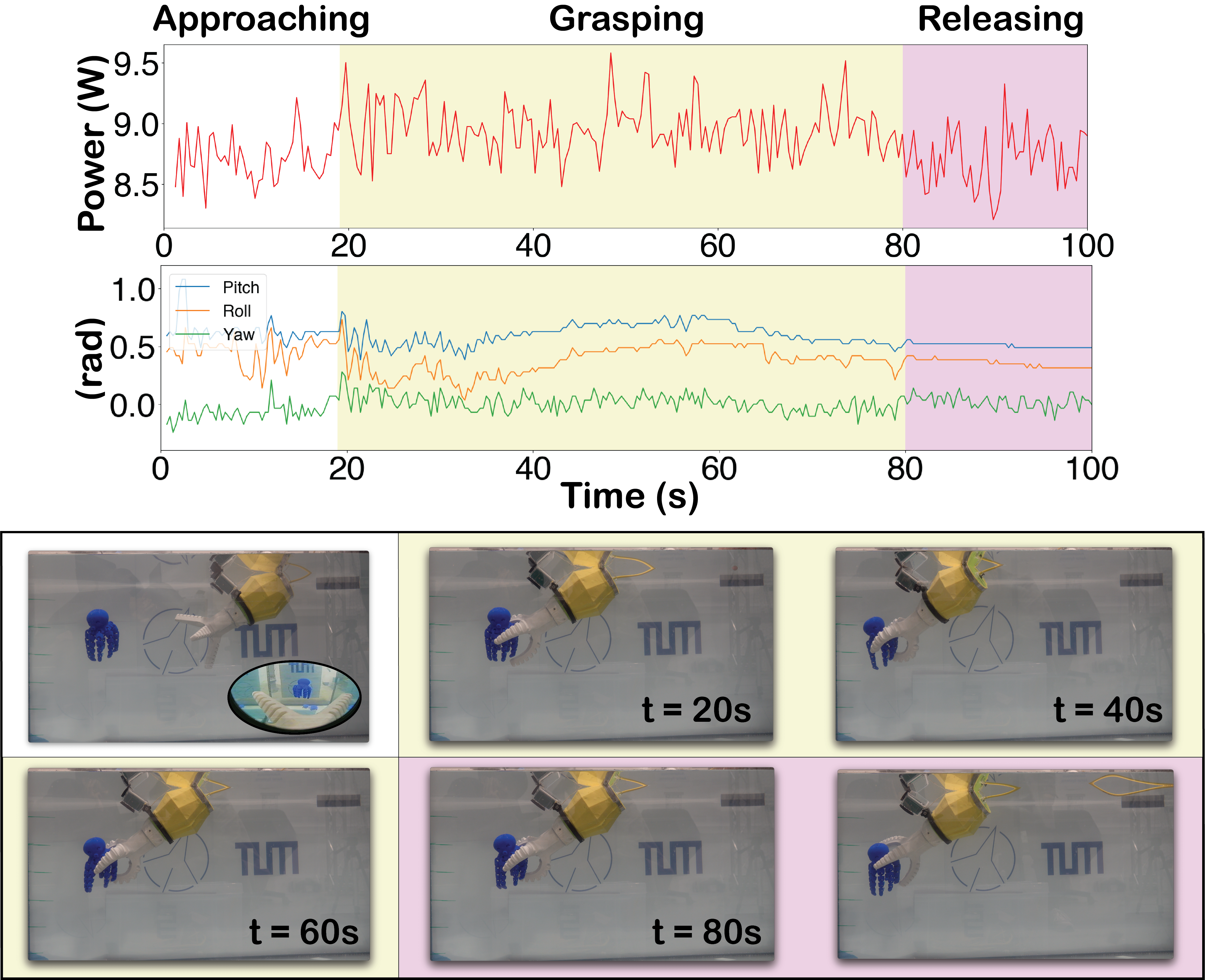}
    \caption{Underwater pod's attitude and energy consumption during the grasping maneuver.}\label{fig:pod_perf}
\end{figure}

The pod's performance was tested by tracking its attitude and power consumption during a typical grasping maneuver. Results are shown in \cref{fig:pod_perf}.
The attitude plot shows that the system remains stable during the entire operation, including the approaching, grasping and releasing phases.
The maximum power consumption is 17 W (5W each for the thrust pumps and grasping pump, 1W each for the pitching pumps) and during these tests the consumed power averages below 9 W with no grasping, and reaches up to 9.5 W when grasping is activated. This is extremely low compared with most of the underwater manipulators, whose maximum power consumption is typically over 30 W without considering underwater locomotion \cite{doi:10.1089/soro.2019.0087}.

\subsection{Field tests}
\begin{figure}[htbp]
    \centering
	\includegraphics[width=\columnwidth]{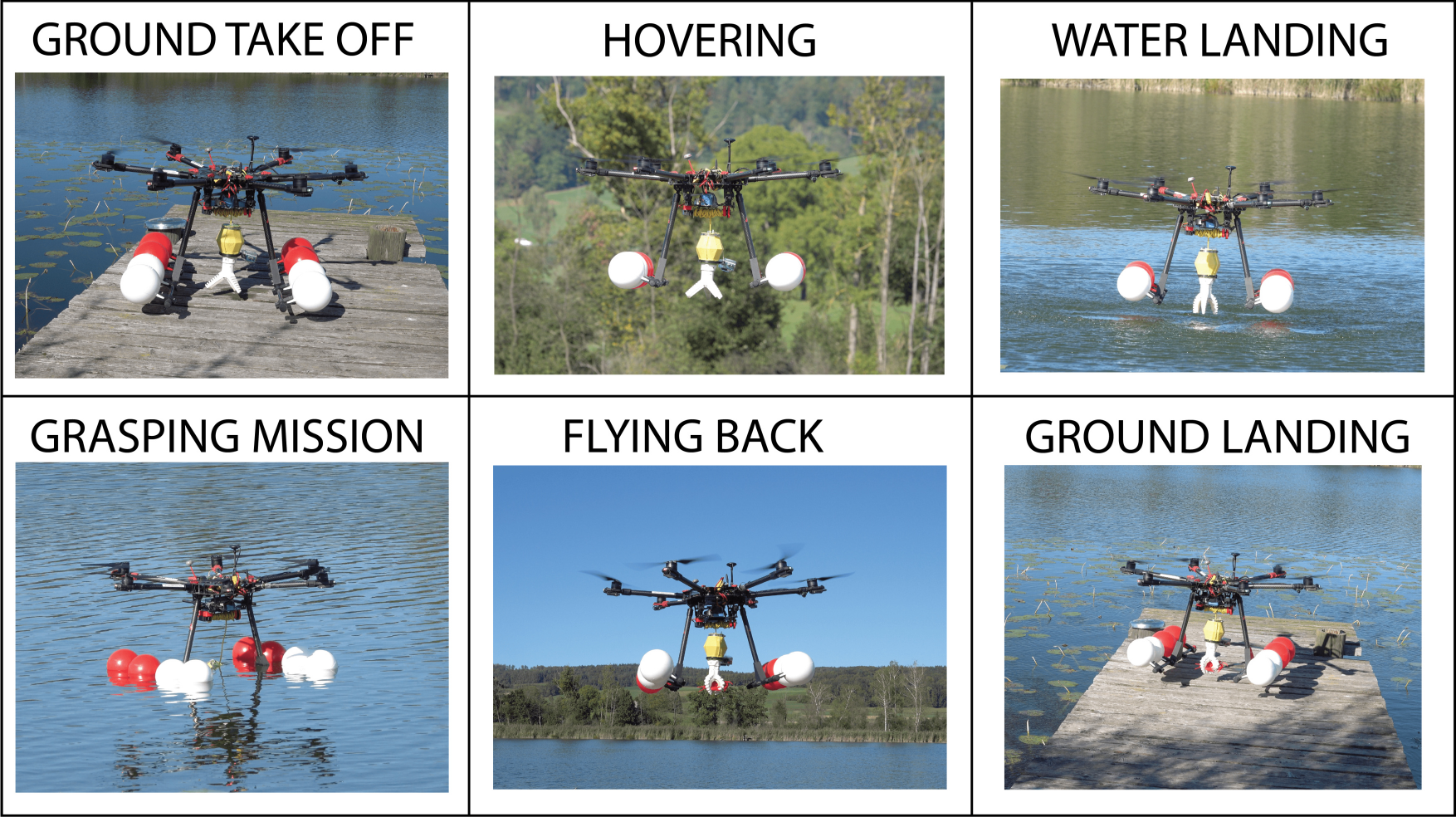}
    \caption{Main mission phases covered in outdoor field tests.}\label{fig:overall_mission}
\end{figure}

The system's aerial deployment validation involved attaching the pod-gripper to a hexacopter for a mission in a lake. Buoy systems enabled the hexacopter to float on the water surface. The mission, as depicted in \cref{fig:overall_mission}, encompassed ground takeoff, flight to the target area, water landing, grasping operation, and takeoff from water to return to the ground.

The hexacopter in use, which incorporates a Tarrot x6 frame and an E1200 DJI propulsion system, has a total weight of 7 kg, inclusive of buoys and batteries. It is capable of exerting a maximum thrust of 9.8 kg, yielding a maximum payload capacity of 2.8 kg. This payload capacity is twice the requirement for transporting the pod-gripper system, as referenced in Chapter \cref{chap:masses}. This allows for the transportation of an additional payload exceeding 800 grams, which corresponds to the maximum dry mass that the gripper can handle, as detailed in \cref{chap:gripper_test}. Furthermore, the winding system effectively maintains close proximity between the gripper system and the drone, as illustrated in the hovering phase in \cref{fig:overall_mission}. This arrangement prevents any pendulum effect between the two and ensures a swift mission execution.

\section{Conclusions and future work}

In this work we design and develop a stand-alone underwater soft gripper attached on a submarine pod, to be deployed on a multi-rotor drone. This system resulted to be capable to accomplish a grasping mission with a force of up to 8 N and reach a water depth of up to 6 meters with the current buoyancy control system.
The gripper is made of silicone with a shape of three-tapered-finger. While the submarine is equipped with differential thrust, pitching attitude control and shape-morphing buoyancy control, which can allow a total volume change of 3.5\%, as well as sensing capabilities with onboard IMU.
This system has been tested deployed on a hexacopter in outdoor tests. The grasping force has been examined with different shapes and dimensions to determine the reliability of the 3-finger-approach. The pod's attitude and energy consumption has been tracked in order to evaluate its performance in terms of underwater locomotion. Not only is the pod-gripper more energy-efficient than typical underwater manipulators, but also the full framework, including aerial deployment, is significantly more efficient than approaches involving deployment via water surface vessels.
One step to improve the control of this setup can be reached by feeding back the measured attitude in order to better control the Yaw, Pitch.
Moreover, the design of the pod can be improved so as to increase the maximum water depth, currently 6 m. 
The gripper's closure time, which currently stands at around 20 seconds, should be enhanced in future endeavors. This can be achieved by reducing the volume within the cavities or, in the case of upscaling the gripper, employing a more powerful pump. 
Moreover, new underwater locomotion methods could be investigated for the pod, for instance the use use of soft fins, to further reduce environmental disturbance.
This concept is part of a broader ecological mission vision in which a fleet of drones can be deployed with interchangeable pods designed for various purposes, such as sampling, grasping, or different types of sensing. These pods can be swapped as needed to adapt to the specific mission requirements, offering a flexible and versatile approach to ecological research and monitoring.



\section*{Acknowledgment}
The authors would like to acknowledge and thank Oscar Pang for his help in testing the system and Hai Pham Nguyen for his help reviewing the manuscript.

\printbibliography[]

@book{zufferey2022between,
  title={Between Sea and Sky: Aerial Aquatic Locomotion in Miniature Robots},
  author={Zufferey, R. and Siddall, Robert and Armanini, Sophie F and Kovac, Mirko},
  year={2022},
  publisher={Springer}
}

@article{shintake2018soft,
  title={Soft robotic grippers},
  author={Shintake, Jun and Cacucciolo, Vito and Floreano, Dario and Shea, Herbert},
  journal={Advanced materials},
  volume={30},
  number={29},
  pages={1707035},
  year={2018},
  publisher={Wiley Online Library}
}

@article{debruyn2020medusa,
  title={Medusa: A multi-environment dual-robot for underwater sample acquisition},
  author={Debruyn, D. and Zufferey, Raphael and Armanini, Sophie F and Winston, Crystal and Farinha, Andr{\'e} and Jin, Yufei and Kovac, Mirko},
  journal={IEEE Robotics and Automation Letters},
  year={2020},
}

@article{farinha2022off,
  title={Off-shore and underwater sampling of aquatic environments with the aerial-aquatic drone MEDUSA},
  author={Farinha, Andr{\'e} Tristany and di Tria, Julien and Reyes, Marta and Rosas, Constanca and Pang, Oscar and Zufferey, Raphael and Pomati, Francesco and Kovac, Mirko},
  journal={Frontiers in Environmental Science},
  volume={10},
  pages={2305},
  year={2022},
  publisher={Frontiers}
}

@article{phillips2018dexterous,
  title={A dexterous, glove-based teleoperable low-power soft robotic arm for delicate deep-sea biological exploration},
  author={Phillips, Brennan T and others},
  journal={Scientific reports},
  volume={8},
  number={1},
  pages={14779},
  year={2018},
  publisher={Nature Publishing Group UK London}
}

@article{galloway2016soft,
  title={Soft robotic grippers for biological sampling on deep reefs},
  author={Galloway, Kevin C and Becker, Kaitlyn P and Phillips, Brennan and Kirby, Jordan and Licht, Stephen and Tchernov, Dan and Wood, Robert J and Gruber, David F},
  journal={Soft robotics},
  volume={3},
  number={1},
  pages={23--33},
  year={2016},
  publisher={Mary Ann Liebert, Inc. 140 Huguenot Street, 3rd Floor New Rochelle, NY 10801 USA}
}

@misc{marchese2014recipe,
  title={A recipe for soft fluidic elastomer robots. Soft robotics, 2 (1), 7-25},
  author={Marchese, AD et al.},
  year={2014}
}

@article{shit2013review,
  title={A review on silicone rubber},
  author={Shit, Subhas C and Shah, Pathik},
  journal={National academy science letters},
  volume={36},
  number={4},
  pages={355--365},
  year={2013},
  publisher={Springer}
}

@article{dollar2005towards,
  title={Towards grasping in unstructured environments: Grasper compliance and configuration optimization},
  author={Dollar, Aaron M and Howe, Robert D},
  journal={Advanced Robotics},
  volume={19},
  number={5},
  pages={523--543},
  year={2005},
  publisher={Taylor \& Francis}
}

@book{laschi2017soft,
  title={Soft robotics: trends, applications and challenges},
  author={Laschi, Cecilia and Rossiter, Jonathan and Iida, Fumiya and Cianchetti, Matteo and Margheri, Laura},
  volume={17},
  year={2017},
  publisher={Springer}
}

@article{hritwick2018orumbot,
  title={OrumBot: Origami-based Deformable Robot Inspired By An Umbrella Structure},
  author={Hritwick  et al.},
  journal={International Conference on Robotics and Biomimetics},
  year={2018}
}

@article{ming2020grasping,
  title={Grasping Marine Products With Hybrid-Driven Underwater Vehicle-Manipulator System},
  author={M. Cai et al},
  journal={TRANSACTIONS ON AUTOMATION SCIENCE AND ENGINEERING},
  volume={17},
  pages={1443--1454},
  year={2020}
}

@article{plum2020sauv,
  title={SAUV—A Bio-Inspired Soft-Robotic Autonomous Underwater Vehicle},
  author={F. Plum et al.},
  journal={frontiers in Neurorobotics},
  volume={14},
  year={2020}
}

@article{katzschmann2018robofish,
  title={Exploration of underwater life with an acoustically controlled soft robotic fish},
  author={R. Katzschmann et al.},
  journal={Science Robotics},
  volume={3},
  year={2018}
}

@article{stoeffler2023jet,
  title={A Combined Rigid-Soft Thruster Based on Jetting Propulsion},
  author={C. Stoeffler et al.},
  journal={International Conference on Soft Robotics (RoboSoft)},
  year={2023},
  publisher={IEEE}
}

@INPROCEEDINGS{10122042,
  author={Mbakop, Steeve and Tagne, Gilles and Lagache, Alice and Youcef-Toumi, Kamal and Merzouki, Rochdi},
  booktitle={2023 IEEE International Conference on Soft Robotics (RoboSoft)}, 
  title={Integrated design of a bio-inspired soft gripper for mushrooms harvesting}, 
  year={2023},
  volume={},
  number={},
  pages={1-6},}

@INPROCEEDINGS{6907203,
  author={Bemfica, J. R. and Melchiorri, C. and Moriello, L. and Palli, G. and Scarcia, U.},
  booktitle={2014 IEEE International Conference on Robotics and Automation (ICRA)}, 
  title={A three-fingered cable-driven gripper for underwater applications}, 
  year={2014},
  volume={},
  number={},
  pages={2469-2474},}

@article{doi:10.1089/soro.2019.0087,
author = {Shen, Zhong and Zhong, Hua and Xu, Erchao and Zhang, Runzhi and Yip, Ki Chun and Chan, Lawrence Long and Chan, Leo Lai and Pan, Jia and Wang, Wenping and Wang, Zheng},
title = {An Underwater Robotic Manipulator with Soft Bladders and Compact Depth-Independent Actuation},
journal = {Soft Robotics},
volume = {7},
number = {5},
pages = {535-549},
year = {2020},
}

@article{doi:10.1089/soro.2022.0215,
author = {Ji, Hui and Lan, Yu and Nie, Songlin and Huo, Linfeng and Yin, Fanglong and Hong, Ruidong},
title = {Development of an Anthropomorphic Soft Manipulator with Rigid-Flexible Coupling for Underwater Adaptive Grasping},
journal = {Soft Robotics},
}

@ARTICLE{9216533,
  author={Liu, J. and Iacoponi, Saverio and Laschi, Cecilia and Wen, Li and Calisti, Marcello},
  journal={IEEE Robotics \& Automation Magazine}, 
  title={Underwater Mobile Manipulation: A Soft Arm on a Benthic Legged Robot}, 
  year={2020},
  volume={27},
  number={4},
  pages={12-26},}

@article{doi:10.1089/soro.2017.0097,
author = {Kurumaya, Shunichi and Phillips, Brennan T. and Becker, Kaitlyn P. and Rosen, Michelle H. and Gruber, David F. and Galloway, Kevin C. and Suzumori, Koichi and Wood, Robert J.},
title = {A Modular Soft Robotic Wrist for Underwater Manipulation},
journal = {Soft Robotics},
volume = {5},
number = {4},
pages = {399-409},
year = {2018},
}

@article{doi:10.1177/0278364920917203,
author = {Zheyuan Gong and Xi Fang and Xingyu Chen and Jiahui Cheng and Zhexin Xie and Jiaqi Liu and Bohan Chen and Hui Yang and Shihan Kong and Yufei Hao and Tianmiao Wang and Junzhi Yu and Li Wen},
title ={A soft manipulator for efficient delicate grasping in shallow water: Modeling, control, and real-world experiments},

journal = {The International Journal of Robotics Research},
volume = {40},
number = {1},
pages = {449-469},
year = {2021},
}

@ARTICLE{6907847,
  author={Stuart, Hannah S. and Wang, Shiquan and Cutkosky, Mark R.},
  journal={IEEE Transactions on Robotics}, 
  title={Tunable Contact Conditions and Grasp Hydrodynamics Using Gentle Fingertip Suction}, 
  year={2019},
  volume={35},
  number={2},
  pages={295-306},
  doi={10.1109/TRO.2018.2880094}}

@article{casalino,
author = {Casalino, Giuseppe and Caccia, Massimo and Caselli, Stefano and Melchiorri, Claudio and Antonelli, Gianluca and Caiti, Andrea and Indiveri, Giovanni and Cannata, G. and Simetti, Enrico and Torelli, Sandro and Sperindè, Alessandro and Wanderlingh, Francesco and Muscolo, Giovanni and Bibuli, Marco and Bruzzone, Gabriele and Zereik, Enrica and Odetti, Angelo and Spirandelli, Edoardo and Ranieri, Andrea and Cataldi, Elisabetta},
year = {2016},
month = {07},
pages = {98-107},
title = {Underwater Intervention Robotics: An Outline of the Italian National Project MARIS},
volume = {50},
journal = {Marine Technology Society Journal},
doi = {10.4031/MTSJ.50.4.7}
}

@INPROCEEDINGS{8593604,
  author={Youakim, Dina and Dornbush, Andrew and Likhachev, Maxim and Ridao, Pere},
  booktitle={2018 IEEE/RSJ International Conference on Intelligent Robots and Systems (IROS)}, 
  title={Motion Planning for an Underwater Mobile Manipulator by Exploiting Loose Coupling}, 
  year={2018},
  volume={},
  number={},
  pages={7164-7171},}

@ARTICLE{8027199,
  author={Simetti, Enrico and Wanderlingh, Francesco and Torelli, Sandro and Bibuli, Marco and Odetti, Angelo and Bruzzone, Gabriele and Rizzini, Dario Lodi and Aleotti, Jacopo and Palli, Gianluca and Moriello, Lorenzo and Scarcia, Umberto},
  journal={IEEE Journal of Oceanic Engineering}, 
  title={Autonomous Underwater Intervention: Experimental Results of the MARIS Project}, 
  year={2018},
  volume={43},
  number={3},
  pages={620-639},
  doi={10.1109/JOE.2017.2733878}}

@ARTICLE{FingersPrecision,
  author={Feix, Thomas and Bullock, Ian M. and Gloumakov, Yuri and Dollar, Aaron M.},
  journal={IEEE Transactions on Haptics}, 
  title={Effect of Number of Digits on Human Precision Manipulation Workspaces}, 
  year={2021},
  volume={14},
  number={1},
  pages={68-82},}

@article{pagoli2021review,
  title={Review of soft fluidic actuators: Classification and materials modeling analysis},
  author={Pagoli, Amir and Chapelle, Fr{\'e}d{\'e}ric and Corrales-Ramon, Juan-Antonio and Mezouar, Youcef and Lapusta, Yuri},
  journal={Smart Materials and Structures},
  volume={31},
  number={1},
  pages={013001},
  year={2021},
  publisher={IOP Publishing}
}

@article{ultragentle,
author = {Nina R. Sinatra  and Clark B. Teeple  and Daniel M. Vogt  and Kevin Kit Parker  and David F. Gruber  and Robert J. Wood },
title = {Ultragentle manipulation of delicate structures using a soft robotic gripper},
journal = {Science Robotics},
year = {2019}}

\end{document}